\documentclass[letterpaper, 10 pt, conference]{ieeeconf} 

\IEEEoverridecommandlockouts                              

\usepackage{graphicx} 
\usepackage{subcaption}
\usepackage{mathptmx} 
\usepackage{amsmath} 
\usepackage{amssymb} 

\title{\LARGE \bf
Task-Space Clustering for Mobile Manipulator Task Sequencing
}

\author{Quang-Nam Nguyen$^{1}$, Nicholas Adrian$^{2}$, and Quang-Cuong Pham$^{1,2,3}$%
\thanks{$^{1}$Singapore Centre for 3D Printing (SC3DP), Nanyang Technological University (NTU), Singapore 
        {\tt\small nam.ngquang@gmail.com}}%
\thanks{$^{2}$HP-NTU Digital Manufacturing Corporate Lab, Nanyang Technological University (NTU), Singapore}%
\thanks{$^{3}$Eureka Robotics, Singapore}%
}

\newtheorem{definition}{Definition}[section]
\newtheorem{remark}{Remark}[section]

\begin{document}

\maketitle
\thispagestyle{empty}
\pagestyle{empty}

\begin{abstract}
        Mobile manipulators have gained attention for the potential in performing large-scale tasks 
        which are beyond the reach of fixed-base manipulators. 
        The Robotic Task Sequencing Problem for mobile manipulators often requires optimizing the motion 
        sequence of the robot to visit multiple targets while reducing the number of base placements. 
        A two-step approach to this problem is clustering the task-space into clusters of targets before 
        sequencing the robot motion. 
        In this paper, we propose a task-space clustering method which formulates the clustering 
        step as a Set Cover Problem using bipartite graph and reachability analysis, then solves it 
        to obtain the minimum number of target clusters with corresponding base placements. 
        We demonstrated the practical usage of our method in a mobile drilling experiment containing 
        hundreds of targets. 
        Multiple simulations were conducted to benchmark the algorithm and also showed that our proposed 
        method found, in practical time, better solutions than the existing state-of-the-art methods.
\end{abstract}

\section{INTRODUCTION}
 
In many robotic applications, a manipulator is required to perform multiple repetitive tasks such as 
drilling, spot-welding, inspection scanning, picking and placing, etc. 
These tasks often involve visiting numerous targets in the workspace. 
The problem of optimizing the robot motion sequence to visit multiple targets is called the Robotic 
Task Sequencing Problem (RTSP) \cite{ref:suarez2018robotsp, ref:wong2020novel}. 

\begin{figure}[tb]
        \centering
        \begin{subfigure}{0.48\linewidth}
                \includegraphics[width=1\linewidth]{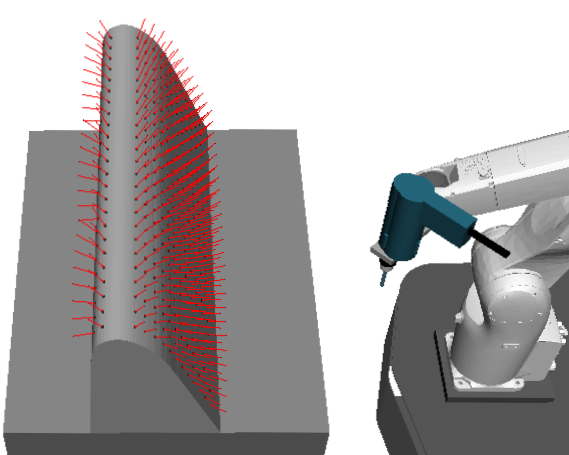}
                \caption{Simulation environment.}
        \end{subfigure}
        \hfill
        \begin{subfigure}{0.48\linewidth}
                \includegraphics[width=1\linewidth]{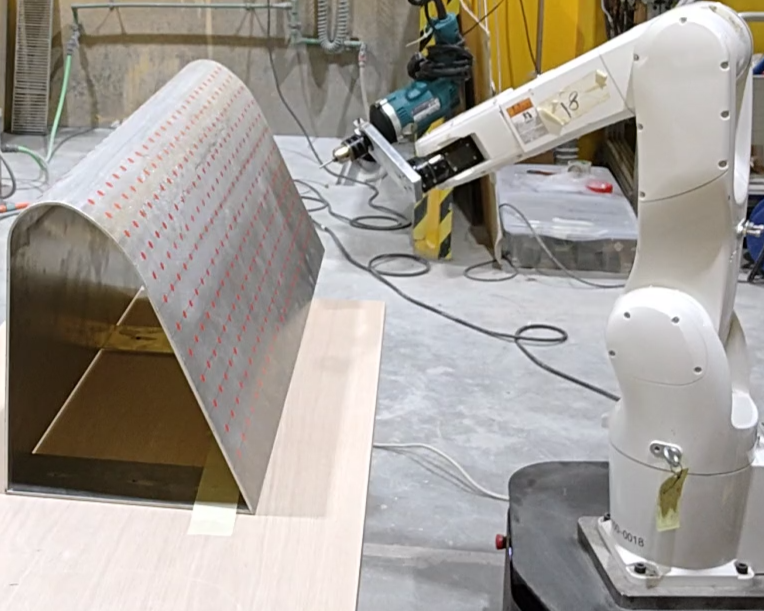}
                \caption{Experimental setup.}
        \end{subfigure}
        \caption{A mobile drilling task containing 336 targets with polar angles between 
        $[110^\circ,150^\circ]$ and azimuthal angles between $[-37^\circ,37^\circ]$ 
        (for 288 front targets) and $[168^\circ,192^\circ]$ (for 48 back targets), 
        on both sides of a $1 m$-long workpiece.}
        \label{fig:demo_task}
\end{figure}

A mobile manipulator is often formed by mounting a manipulator, such as a robotic arm, on a mobile base, 
which helps extend the workspace of the manipulator \cite{ref:vafadar2018optimal, ref:xu2021planning}. 
In mobile manipulation tasks, the targets are often distributed in a large workspace that is beyond 
the reach of fixed-base manipulators, such as drilling on both sides of a workpiece as shown in 
Fig. \ref{fig:demo_task}, picking and placing \cite{ref:xu2021planning}, surface inspection 
\cite{ref:malhan2022finding}, concrete 3D printing \cite{ref:tiryaki2019printing}, etc. 

Mobile Manipulator RTSP has extra complexities due to the added degrees-of-freedom (DOFs) from the 
mobile base. 
Moreover, minimizing the number of base placements often has higher priority over the arm motion 
because of the higher localization uncertainty of the mobile base, especially during accelerating from 
and to a placement. 
Therefore, we propose solving the Mobile Manipulator RTSP in 2 steps: 
\begin{itemize}
        \item Task-space clustering: dividing all targets into a minimum number of clusters to cover 
        the whole workspace. 
        \item Task sequencing: finding the optimal motion sequence for the robot to visit all targets 
        in the shortest path. 
\end{itemize}

Our main contribution in this paper is a task-space clustering method to divide all targets into a 
minimum number of clusters, for each of which there is a base pose where the manipulator can reach 
all the inside targets. 
The key of our method is the formulation of a Set Cover Problem (SCP) by constructing a bipartite 
graph (bigraph) between task-space targets and floor points using reachability analysis. 
The method has been integrated with task sequencing in a planning algorithm used for a mobile drilling 
experiment. 

The remainder of this paper is organized as follows. 
In the next section, we discuss related works in solving RTSP for fixed-base and mobile manipulators. 
In section III, we describe the task and introduce the pipeline of our method which will be discussed 
in more details in sections IV, V. 
In section VI, we present the experimental results, benchmarking, and comparison with existing methods. 

\section{RELATED WORKS}

Several works have been done on solving RTSP for fixed-base robots. 
A recent survey can be found in \cite{ref:alatartsev2015robotic}. 
To the best of our understanding, two of the fastest near-optimal RTSP solvers are RoboTSP 
\cite{ref:suarez2018robotsp} and Cluster-RTSP \cite{ref:wong2020novel} which use different approaches. 
They reported the planning time of few minutes for a drilling task with hundreds of targets. 

RTSP has also been solved for mobile manipulators such as in \cite{ref:zacharia2005optimal} where 
authors conducted a search on 10-DOF configuration-space (C-space) to minimize overall configuration 
displacement based on genetic algorithms. 
An approach to extend the usage of RTSP solvers for mobile manipulators is finding optimal base 
placements around the workspace based on robot's reachability or manipulability without targets' 
information, but these placements may not be optimal for every task. 
In \cite{ref:zacharias2007capturing}, reachability map was introduced as a 3D workspace representation 
that represents the reachability probability of points in task-space. 
The reachability map was then used in \cite{ref:zacharias2008positioning} to find robot placements for 
a constrained linear trajectory and was extended for 3D trajectories in \cite{ref:zacharias2009using}. 
In \cite{ref:vahrenkamp2012manipulability}, extended manipulability measure is used as precomputed 
workspace representation instead of reachability. 
In \cite{ref:vahrenkamp2013robot}, robot placements to reach a target with probabilities can be found 
using Inverse Reachability Distribution.

Some works have also attempted to solve the Mobile Manipulator RTSP using the 2-step approach: 
clustering the workspace before sequencing. 
In \cite{ref:xu2021planning}, a mobile manipulator performed picking multiple objects from multiple 
trays. The problem of finding minimum base placements to visit these trays was formulated as the 0-1 
knapsack problem and solved in 139 seconds for 22 trays. 
However, the initial clusters like the trays are usually not available in other cases. 
Meanwhile, our method focuses on tasks with more general settings where only the position and 
orientation of the targets are given. 
In \cite{ref:malhan2022finding}, a mobile manipulator performed inspection scanning with a camera 
attached to its end-effector. 
The authors proposed finding reachable targets for each point on the floor then using a search-based 
algorithm to find the optimal base sequence, and reported that their method produces better results 
than other existing methods. 
In contrast, our method finds reachable floor points for each target and formulates the problem of 
finding optimal base poses as a Set Cover Problem which can be solved by near-optimal solvers.

\section{METHODOLOGY OVERVIEW}

\subsection{Task description}

\subsubsection{Targets}
We focus on 5-dimensional targets (3D position and 2D orientation) such as in drilling, milling, 
welding tasks. 
Consider a 5D target, we use the Cartesian coordinate system to locate the position of the target 
while the orientation is represented by a unit vector in the spherical coordinate system. 
A set of 5D targets is given in world frame as: 
\begin{equation}
        T = \{\mathbf{t}_1, \dots, \mathbf{t}_n\}, \quad \mathbf{t}_i = (x^t_i, y^t_i, z^t_i, \theta^t_i, \phi^t_i)
        \label{eq:targets}
\end{equation}
where $\theta^t_i$ and $\phi^t_i$ denote polar angle and azimuthal angle, respectively, of the 
target's orientation vector in the spherical coordinate system. 
Since the 6\textsuperscript{th} dimension, such as rotation about the drilling direction, is irrelevant, 
one can set specific values for it as tuning parameters, similarly to \cite{ref:suarez2018robotsp}.  

\subsubsection{Floor points} 
The 2D floor where the mobile base is moving on can be discretized into a grid of discrete points: 
\begin{equation}
        F = \{ \mathbf{f}_1, \dots, \mathbf{f}_m \}, \quad \mathbf{f}_j = (x^f_j, y^f_j), 
        \label{eq:floor}
\end{equation}

\begin{remark}[Feasible tasks]
\label{rem:feasible_tasks}
        When the mobile base is placed at position $\mathbf{f}_j$ on the floor with free orientation, 
        there is a set $S_j$ of targets that the manipulator can reach. 
        A task is \textit{feasible} if the union of these sets equals the set of all targets: 
        \begin{equation}
                \label{eq:feasible_task}
                \bigcup_{j=1, \dots,m} S_j = T
        \end{equation} 
\end{remark}

\subsubsection{Problem definition}

The goal of RTSP is to find the optimal sequence of robot configurations to visit all targets subjected 
to a certain optimization objective such as motion time \cite{ref:alatartsev2015robotic}. 
Finding the optimal motion time depends on solving for joint velocity and acceleration constraints, 
which is difficult to find in practical time for substantial target count. 
Instead, it will be more tractable to optimize for shortest configuration path as optimization objective. 
Beyond that, as we are solving the Mobile Manipulator RTSP, we also prioritize minimizing the number of 
base placements because of higher localization uncertainty of the mobile base compared to the robotic arm.

\subsection{Proposed method}

To solve the Mobile Manipulator RTSP, we follow a 2-step approach: 
clustering the targets in task-space into a minimum number of clusters while simultaneously determining 
the \textit{corresponding base poses} to reach targets in each cluster, then sequencing the robot 
configurations to visit every target while optimizing the total length of the configuration path. 

We propose the task-space clustering as follows (Fig. \ref{fig:pipeline}): 

\subsubsection{Reachability analysis}
Generate and analyse the kinematic reachability of the robot to find a \textit{geometric reachable region} 
inside which every point is reachable (Fig. \ref{fig:fkr}). 

\subsubsection{Bigraph connection}
We formulate the task-space clustering as a Set Cover Problem \cite{ref:williamson2011design, 
ref:agarwal2014near} using a bigraph which connects elements of the set of targets $T$ with the set of 
floor points $F$ based on the previous reachability analysis. 

\subsubsection{Cluster assignment}
Solve the SCP for the minimum number of clusters needed to cover all targets and the corresponding 
base poses to visit them. 

\begin{figure*}[tb]
        \centering
        \begin{subfigure}{0.3\linewidth}
                \includegraphics[width=1\linewidth]{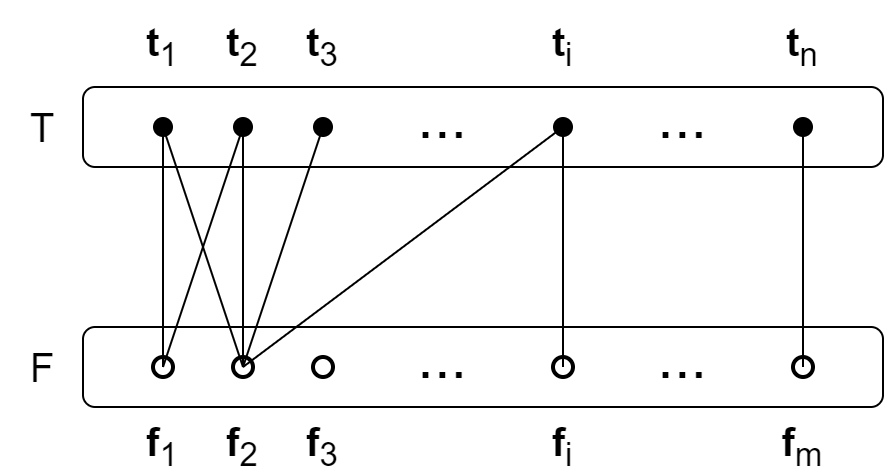}
                \caption{A bigraph connecting sets $T$ and $F$.}
                \label{fig:bigraph}
        \end{subfigure}
        \hfill
        \begin{subfigure}{0.3\linewidth}
                \includegraphics[width=1\linewidth]{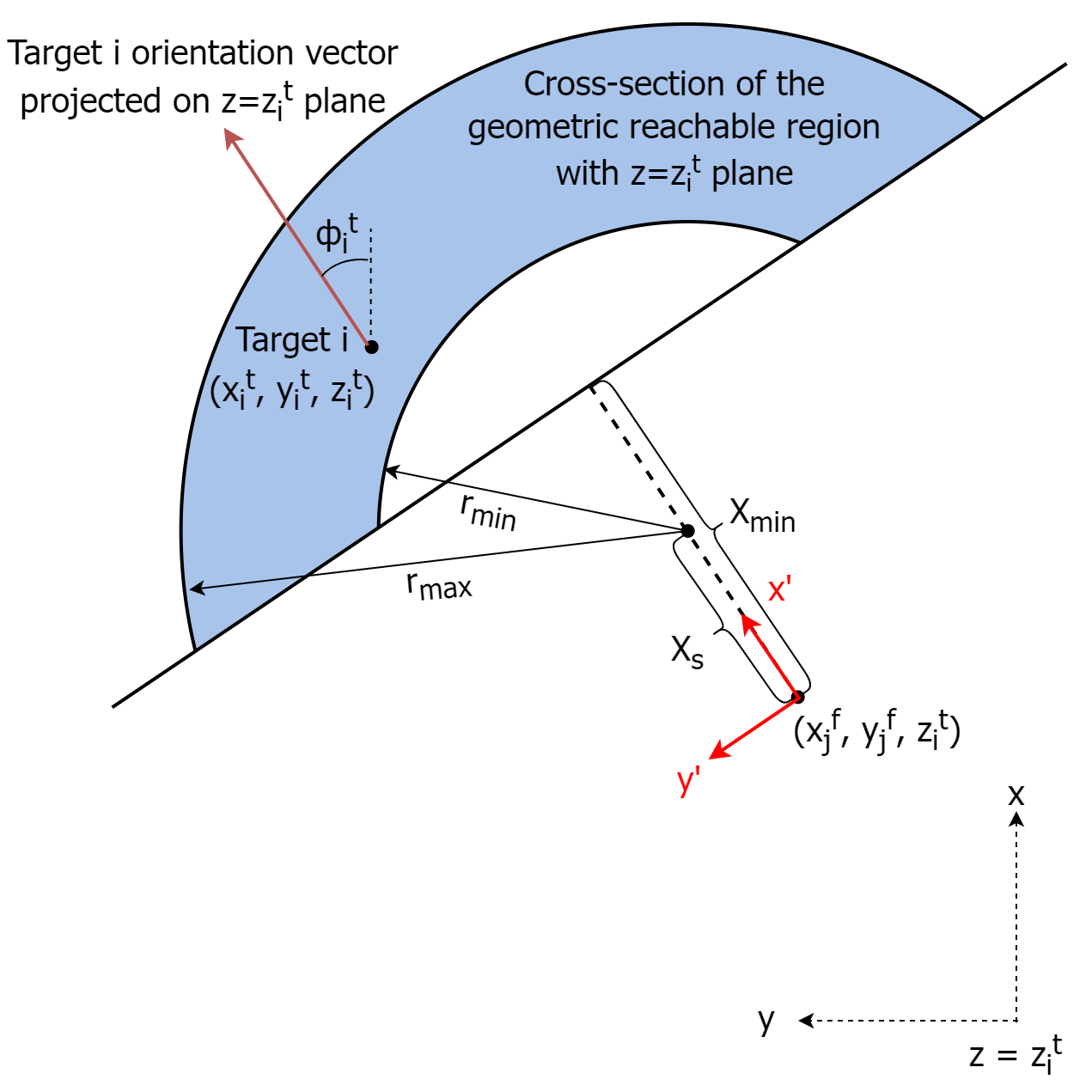}
                \caption{Horizontal plane at $z = z_i^t$.}
                \label{fig:connection1}
        \end{subfigure}
        \hfill
        \begin{subfigure}{0.3\linewidth}
                \includegraphics[width=1\linewidth]{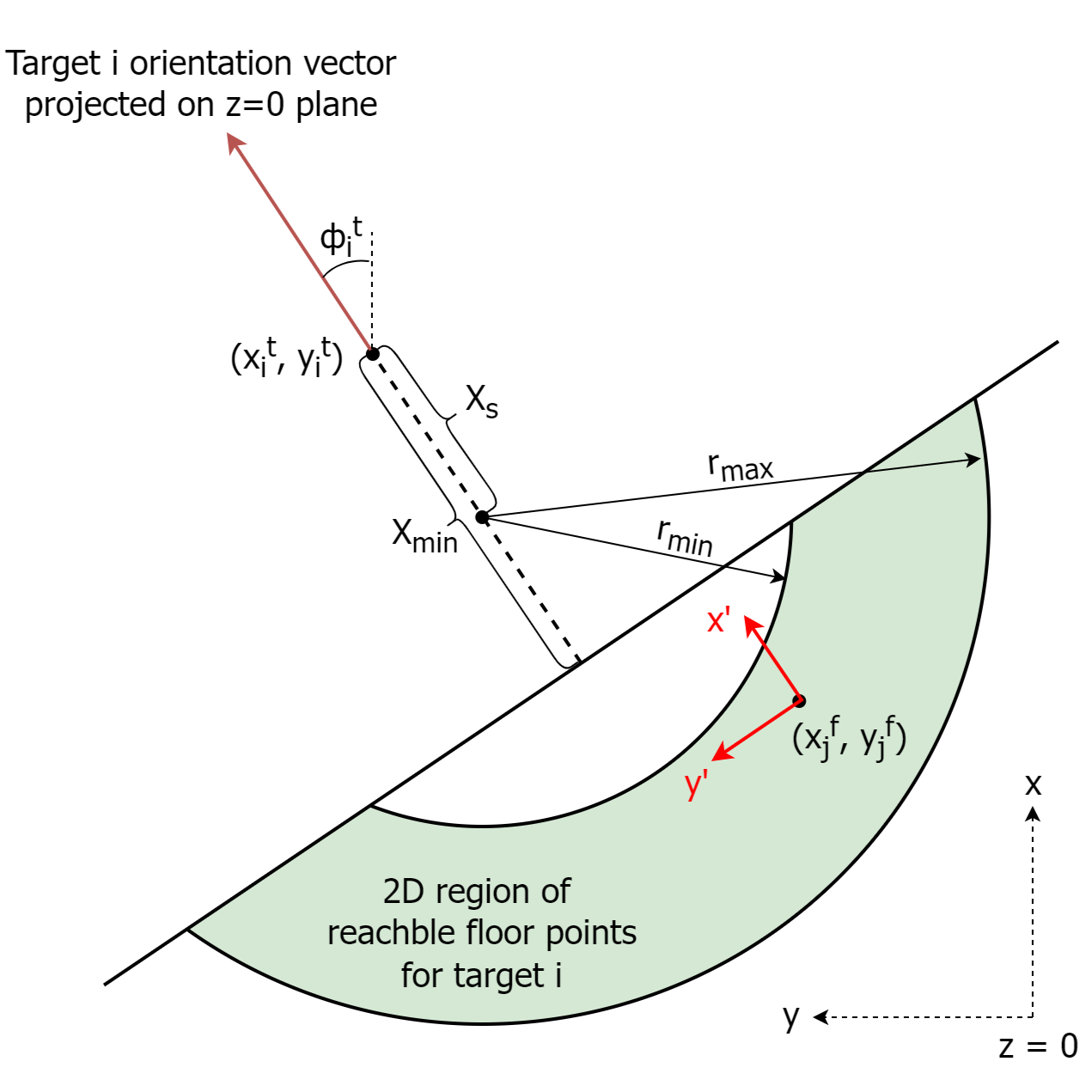}
                \caption{Horizontal plane at $z = 0$.}
                \label{fig:connection2}
        \end{subfigure}
        \hfill
        \begin{subfigure}{0.22\linewidth}
                \includegraphics[width=1\linewidth]{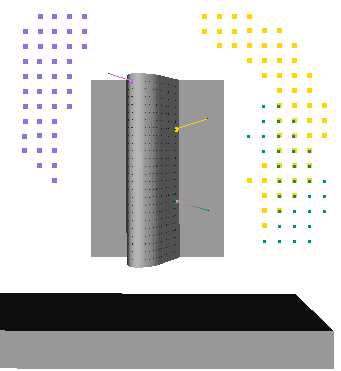}
                \caption{Reachable floor points.}
        \end{subfigure}
        \hspace{0.1\linewidth}
        \begin{subfigure}{0.40\linewidth}
                \includegraphics[width=1\linewidth]{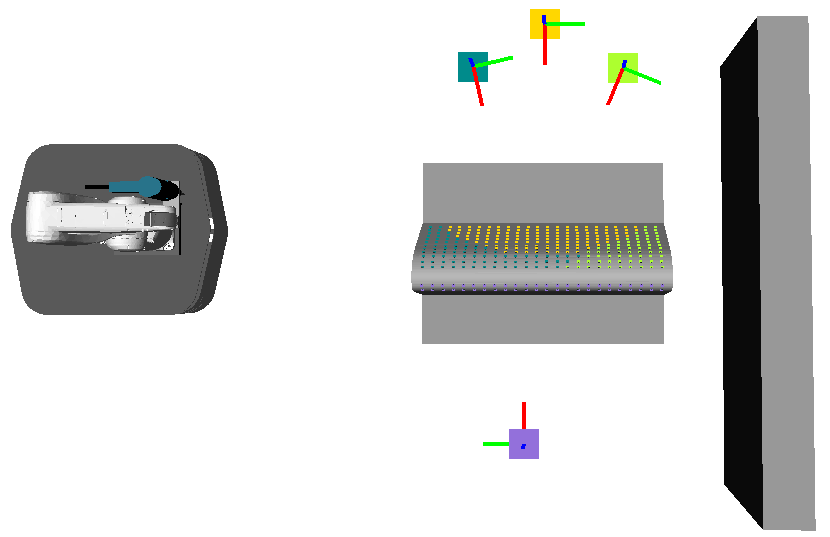}
                \caption{Clusters of targets with corresponding base poses.}
                \label{fig:solution}
        \end{subfigure}
        \caption{The pipeline of our task-space clustering method. 
        (a) Bigraph whose edges connect vertices between set $T$ of targets and set $F$ of floor points. 
        (b) Target $i$ is reachable by the robot placed at point $(x_j^f, y_j^f)$ on the floor if 
        it is inside a cross-section of the geometric reachable region with the $z=z_i^t$ plane. 
        (c) Inversely, target $i$ is kept inside this reachable cross-section if the base is placed 
        inside a respective 2D region on the floor. 
        (d) Example of reachable floor points for some targets at different locations. 
        (e) The minimum set of target clusters with corresponding base poses is found by solving a 
        uniform-cost SCP.}
        \label{fig:pipeline}
\end{figure*}

For the task sequencing step, we implement RoboTSP \cite{ref:suarez2018robotsp} with some modifications:
optimizing Euclidean distance of the base and manipulator paths in task-space, then optimizing C-space 
path length. Details of these clustering and sequencing methods are discussed in sections IV and V.

\section{TASK-SPACE CLUSTERING}

The purpose of our task-space clustering method is to find the minimum number of base poses where the 
robot can reach all targets in the corresponding clusters which cover every target in the task-space. 
We propose formulating this clustering problem as a uniform-cost SCP \cite{ref:williamson2011design} 
as follows: 

\begin{definition}[Uniform-cost SCP]
        \label{def:scp}
        A uniform-cost SCP consists of a set $U = \{u_1, \dots, u_n \}$ often called the \textit{universe} 
        and a \textit{collection} $S = \{s_1, \dots, s_m \}$ of subsets of $U$, that is 
        $s_j \subset U \ \forall j$. The union of these sets must be equal to $U$: 
        \begin{equation}
                \label{eq:union}
                \bigcup_{j=1,\dots,m} s_j = U
        \end{equation}
        The optimization problem is to find the minimum sub-collection $C \subset S$ that still 
        covers all elements of $U$. 
\end{definition}

The universe $U$ and collection $S$ in our robotic case are: 
\begin{itemize}
        \item The universe: $U = \{u_1, \dots, u_n \}$ is the set of indices $u_i = i$ of targets 
        in $T = \{\mathbf{t}_1, \dots, \mathbf{t}_n\}$ which was defined in (\ref{eq:targets}).
        \item Each set $s_j$ contains the indices of targets that are reachable when the robot is 
        placed at point $\mathbf{f}_j$ on the floor $F = \{ \mathbf{f}_1, \dots, \mathbf{f}_m \}$ 
        which was defined in (\ref{eq:floor}). 
\end{itemize}

We propose a fast method to construct the collection $S$ of \textit{reachable sets} 
$S = \{s_1, \dots, s_m \}$ by an inverse approach: for each target, find \textit{reachable floor points} 
where robot can reach the target, based on reachability analysis. 
The condition (\ref{eq:union}) is equivalent to the condition of feasible task 
(Remark \ref{rem:feasible_tasks}).

\subsection{Reachability analysis}
The purpose of this step is to define a reachable region relative to the robot which meets two 
requirements: the robot can reach any point in this region within a specified range of orientations, 
and it has a simple geometric shape. 

To fulfil the first requirement, we construct a \textit{reachability database} which involves 
discretizing the Cartesian space relative to the robot into 3D voxels. 
For each voxel position, we perform Inverse Kinematics (IK) and record all \textit{valid voxels} 
at which IK solutions exist for all sampling orientations within a specified range as follows: 

\begin{itemize}
        \item Sampling polar angles: specify a range that bounds the polar angles of all targets, 
        and discretize it into $N_{sam}$ sampling polar angles: 
        $\theta^{sam}_1, \theta^{sam}_2, \dots, \theta^{sam}_{N_{sam}}$. 
        We take $N_{sam} = 10$ values from $[110^\circ, 150^\circ]$ for our test case. 
        If the polar range is wider, divide it into smaller ranges. 
        \item Sampling azimuthal angles: for reachability database generation, set the sampling 
        azimuthal angle $\phi^{sam} = 0$. 
\end{itemize}

\begin{figure}[tb]
        \centering
        \begin{subfigure}{0.48\linewidth}
                \includegraphics[width=1\linewidth]{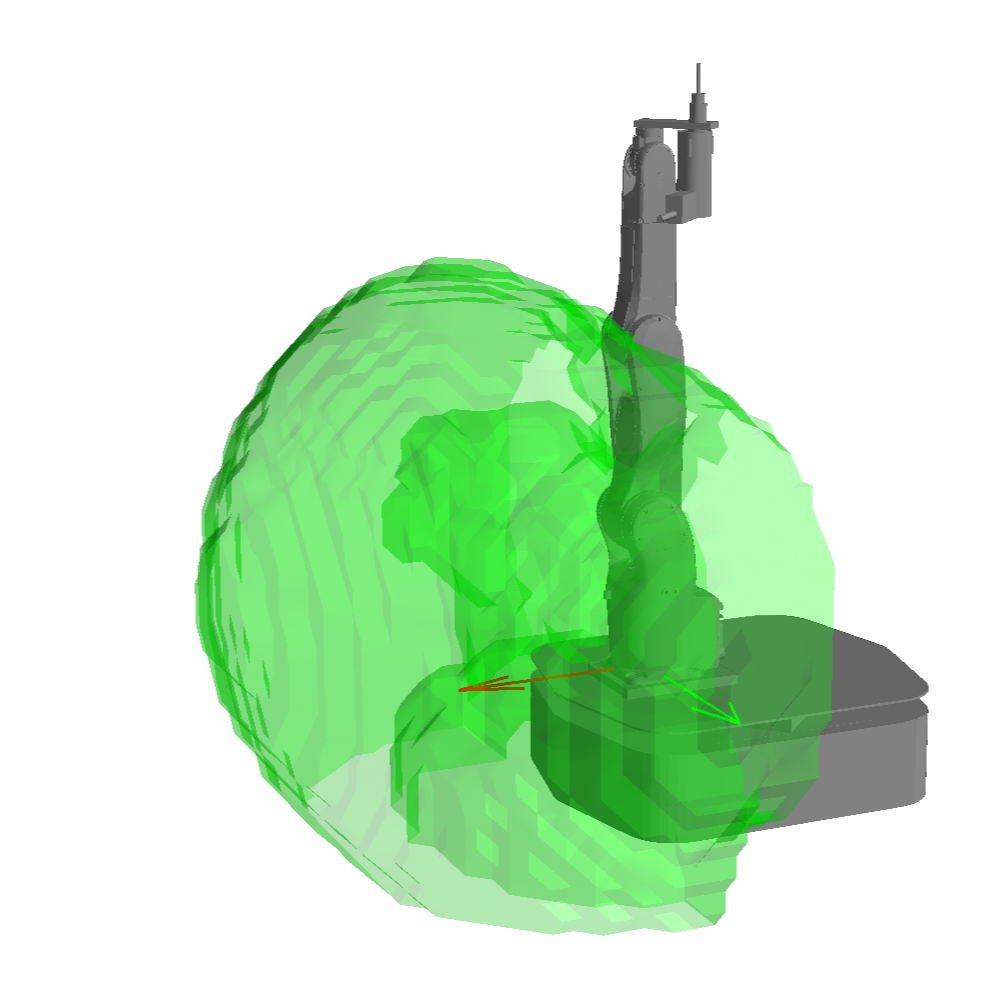}
                \caption{Reachability database.}
                \label{fig:fkr_raw}
        \end{subfigure}
        \hfill
        \begin{subfigure}{0.48\linewidth}
                \includegraphics[width=1\linewidth]{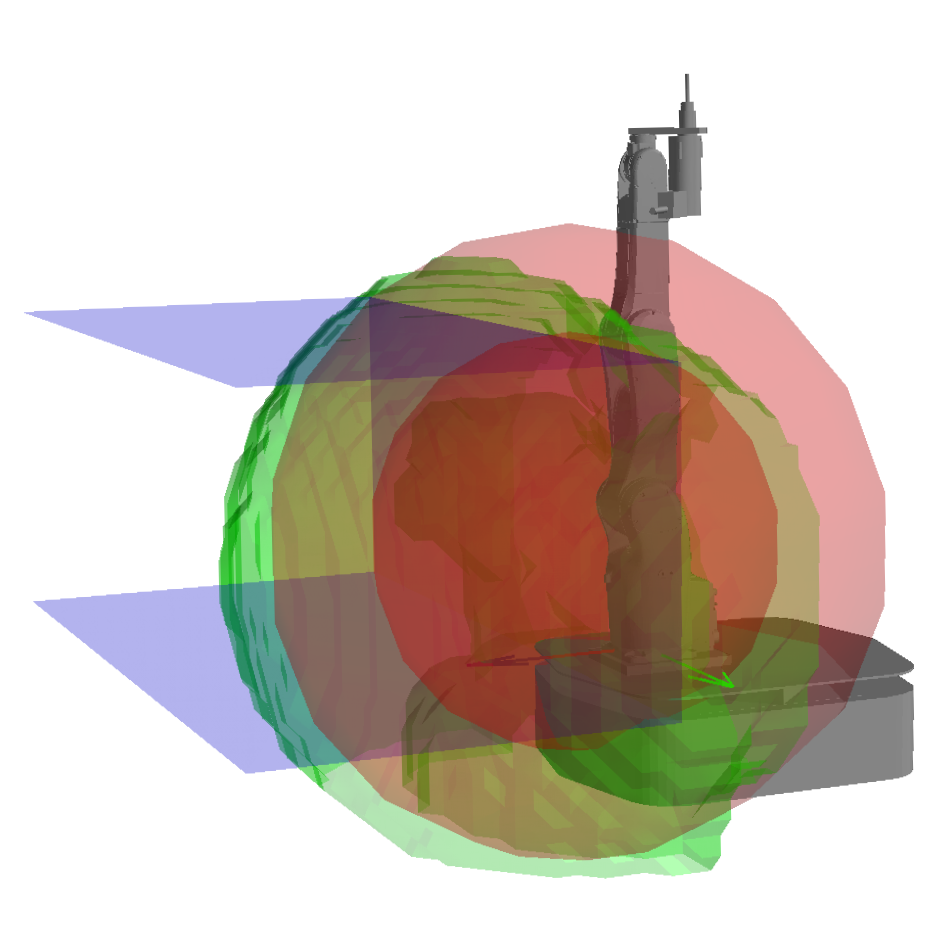}
                \caption{Geometric reachable region.}
                \label{fig:fkr_analysed}
        \end{subfigure}
        \caption{(a) Reachability database: all voxels between the green surfaces are reachable. 
        (b) Geometric reachable region: 3 limit planes and 2 limit spherical surfaces are determined.}
        \label{fig:fkr}
\end{figure}

Figure \ref{fig:fkr_raw} illustrates the cloud of valid voxels after the database generation. 
Note that a reachability database can be used for tasks with targets having polar angles within 
the sampling polar range, while the azimuthal angles can have any values, based on the following 
observation and setting: 

Firstly, we define the \textit{robot frame} having its $z'$-axis along the rotational axis of the first 
joint which is parallel to the ``up'' direction with $z'=0$ at the floor level, while the $x'$-axis and 
$y'$-axis are parallel to the ``front'' and ``left'' directions of the robot, respectively. 
This is because the axis of the first joint is usually vertical after mounted onto the mobile base. 

Secondly, for reachability database generation, we restrict the range of the manipulator's first joint 
to, for example, $[-j_1^{res}, j_1^{res}] = [-90^\circ, 90^\circ]$ which is narrower than our robot's 
hardware limit of $[-j_1^{lim}, j_1^{lim}] = [-170^\circ, 170^\circ]$, since we are only interested in 
the space in front of the robot. 
This setting enables rotating the reachable region about the $z'$-axis by rotating the first joint 
instead of rotating the mobile base, subjected to a \textit{reachable azimuthal width per base pose}: 
\begin{equation}
        \Delta\phi_{max} = 2 \times (j_1^{lim} - j_1^{res})
        \label{eq:azimuthal_width}
\end{equation}
 
To fulfil the second requirement, we find a geometric shape that fits inside the raw valid voxel cloud, 
as follows: 

\subsubsection{Step 1}
Manually specify a region by 3 \textit{limit planes}: 
\begin{itemize}
        \item 2 horizontal planes that bound the $z$-range of the targets: 
        $Z_{min} \leq z' \leq Z_{max}$ with $Z_{min} \leq \min{z^t_i}$, $Z_{max} \geq \max{z^t_i}$. 
        \item 1 vertical plane to keep a safe distance from the robot to avoid potential collisions: 
        $x' \geq X_{min}$ 
\end{itemize}

\subsubsection{Step 2}
Define 2 \textit{limit spherical surfaces} having the same centre at a position $(X_s, 0, Z_s)$ 
relative the robot: 
\begin{equation}
        \begin{aligned} 
                &X_s = l \sin{\overline{\theta}^{sam}} \\
                &Z_s = z'_{j2} + l \cos{\overline{\theta}^{sam}}
        \end{aligned}
\end{equation}
where $\overline{\theta}^{sam}$ is the average value of the sampling polar angles, $l$ is the 
end-effector's length (in our case it is calculated from the second-last joint to the end-effector's tip), 
and $z'_{j2}$ is the $z'$-coordinate of the second joint. 
This is because when no end-effector is attached, the reachable region often has a near-spherical 
shape centred at the average location of the second joint \cite{ref:porges2014reachability}, that is 
$(\overline{x}'_{j2}, \overline{y}'_{j2}, \overline{z}'_{j2}) = (0, 0, z'_{j2})$. 
Note that the $z'$-axis was set along the first joint's rotational axis. 

\subsubsection{Step 3}
We compute the minimum and maximum radii $R_{min}, R_{max}$ corresponding to 2 limit spherical surfaces 
such that all voxels inside a closed region formed by 2 limit spherical surfaces and 3 limit planes are 
valid voxels (see Fig. \ref{fig:fkr_analysed}). This geometric reachable region is defined by: 
\begin{equation}
        \begin{aligned} 
                &Z_{min} \leq z' \leq Z_{max} \\
                &x' \geq X_{min} \\
                &R_{min} \leq \sqrt{(x'-X_s)^2 + y'^2 + (z'-Z_s)^2} \leq R_{max}
        \end{aligned}
        \label{eq:fkr}
\end{equation}
In our test case: $X_{min} = 0.40m$, $Z_{min} = 0.40m$, $Z_{max} = 1.20m$, $X_s = 0.22m$, $Z_s = 0.64m$, 
$R_{min} = 0.51m$, $R_{max} = 0.84m$.

\begin{remark}[Checking reachable targets]
        \label{rem:check}
        In case the target's azimuthal angle is $\phi^t_i = 0$, this target is reachable if its position 
        is inside the geometric reachable region of robot, i.e. all conditions in (\ref{eq:fkr}) are 
        satisfied. In case $\phi^t_i \neq 0$, one can rotate the geometric reachable region by $\phi^t_i$ 
        about the $z'$-axis. 
\end{remark}

\subsection{Bigraph connection}

To construct the reachable sets $s_1, \dots, s_m$ that connect the elements of $T$ with elements of $F$, 
the direct approach is to place the robot at point $\mathbf{f}_j$ on the floor and record to $s_j$ the 
indices of targets such that at least one IK solution is found to reach them. 
However, solving IK in $n \times m$ iterations is time-consuming for numerous targets. 
A faster way of finding reachable targets for each floor point is using Remark \ref{rem:check} which 
also requires $n \times m$ iterations. 

An inverse approach is using the Inverse Reachability Distribution as in \cite{ref:vahrenkamp2013robot} 
or similarly in \cite{ref:malhan2022finding, ref:porges2014reachability} to find reachable base 
placements for each target. 
However, for multiple targets, this approach may require matching the inverted voxels to the grid points 
on the floor for every target. 
We propose another inverse approach which requires $n$ fast iterations: 
based on the geometric reachable region defined in (\ref{eq:fkr}), find reachable floor points for each 
target as follows.

Consider a target $\mathbf{t}_i = (x^t_i, y^t_i, z^t_i, \theta^t_i, \phi^t_i)$, the robot placed at floor 
point $\mathbf{f}_j = (x^f_j, y^f_j)$ with orientation 
$\varphi_j \in [\phi^t_i - \Delta\phi_{max}/2, \ \phi^t_i + \Delta\phi_{max}/2]$ can rotate its geometric 
reachable region by $(\phi^t_i - \varphi_j)$ to check whether it can reach target $i$. 
Substituting $x' = (x^t_i - x^f_j)\cos\phi^t_i+(y^t_i - y^f_j)\sin\phi^t_i, \ 
y' = (y^t_i - y^f_j)\cos\phi^t_i-(x^t_i - x^f_j)\sin\phi^t_i, \ z' = z^t_i$ into (\ref{eq:fkr}), we get: 
\begin{equation}
        \begin{aligned} 
                &(x^t_i - x^f_j)\cos\phi^t_i+(y^t_i - y^f_j)\sin\phi^t_i \geq X_{min}\\
                &(x^t_i - x^f_j - X_s\cos\phi^t_i)^2 + (y^t_i - y^f_j - X_s\sin\phi^t_i)^2 \geq r_{min}^2\\
                &(x^t_i - x^f_j - X_s\cos\phi^t_i)^2 + (y^t_i - y^f_j - X_s\sin\phi^t_i)^2 \leq r_{max}^2\\
                &r_{max} = \sqrt{R_{max}^2 - (z^t_i - Z_s)^2}\\
                &r_{min} = 
                \begin{cases}
                        0, \quad \text{if} \ (z^t_i - Z_s)^2 > R_{min}^2\\
                        \sqrt{R_{min}^2 - (z^t_i - Z_s)^2}, \quad \text{otherwise}
                \end{cases}
        \end{aligned}
        \label{eq:connection}
\end{equation}
Note that the first line in (\ref{eq:fkr}) becomes $Z_{min} \leq z^t_i \leq Z_{max}$ which is always 
satisfied by the definition of $Z_{min}, \ Z_{max}$. 

We use (\ref{eq:connection}) to find all reachable floor points $(x^f_j, y^f_j)$ for each target 
(see Fig. \ref{fig:connection1},\ref{sub@fig:connection2}), which takes $n$ iterations. 
From $n$ obtained sets of reachable floor points, we can easily construct $m$ sets of indices of 
reachable targets for every floor point: $S = \{s_1, \dots, s_m \}$. 
Thus, two sets $T$ and $F$ in the bigraph are connected, and the uniform-cost SCP is obtained.

\subsection{Cluster assignment}
Although the optimization problem of SCP is NP-complete \cite{ref:williamson2011design, ref:agarwal2014near}, 
there are many heuristic methods to find near-optimal solutions to SCP in polynomial time (see VI.B). 
We solve the SCP obtained above to get the minimum number of reachable floor points, called 
\textit{chosen floor points}, with the corresponding \textit{chosen sets} of indices of reachable targets. 

\begin{figure}[tb]
        \centering
        \includegraphics[width=0.4\linewidth]{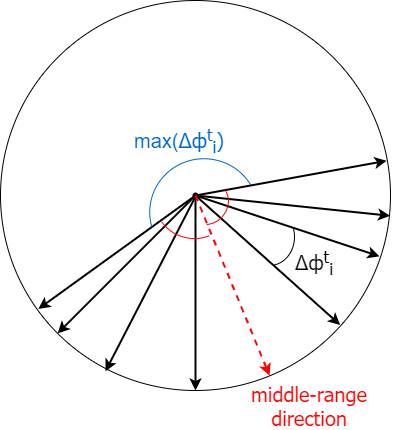}
        \caption{An example of azimuthal angles in a set $s_k$, how to calculate azimuthal range 
        and choose the base orientation.}
        \label{fig:azimuthal}
\end{figure}

Since the range of azimuthal angles in some chosen sets may be wider than the reachable azimuthal width 
per base pose $\Delta\phi_{max}$ defined in (\ref{eq:azimuthal_width}), clusters are assigned as follows: 
\begin{itemize}
        \item For each chosen set $s_k$, sort its elements by their azimuthal angles, then the 
        width of azimuthal range is: 
        \begin{equation}
                \Delta\phi^s_k = 2\pi - \max_{i \in s_k}(\Delta\phi^t_i) 
        \end{equation}
        where $\Delta\phi^t_i$ is the azimuthal difference from target $i$ to the next sorted 
        target (except $\Delta\phi^t_{last} = 2\pi + \phi^t_{first} - \phi^t_{last}$). 
        \item If a chosen set $s_k$ satisfies $\Delta\phi^s_k \leq \Delta\phi_{max}$, assign 
        its elements into a cluster $C_j = s_k$ and set the corresponding base position at 
        $(x^b_j, y^b_j) = (x^f_k, y^f_k)$ with base orientation $\varphi^b_j$ at the middle 
        of $s_k$'s azimuthal range (see Fig. \ref{fig:azimuthal}). 
        \item Update other chosen sets, then repeat two steps above. 
        \item Split an unsatisfied chosen set into smaller sets satisfying 
        $\Delta\phi^s_{k,new} \leq \Delta\phi_{max}$, then repeat the previous steps. 
\end{itemize}

Let $p$ be the number of the clusters obtained from above, then the clusters and their corresponding 
base poses are: 
\begin{equation}
        \begin{aligned} 
        &C = \{C_1, \dots, C_p\} \\
        &B = \{\mathbf{b}_1, \dots, \mathbf{b}_p\}, \quad \mathbf{b}_j = (x^b_j, y^b_j, \varphi^b_j)
        \end{aligned}
\end{equation}

\section{TASK SEQUENCING}

This task sequencing step aims to optimize the configuration path of the robot to visit all targets, 
which is related to the General Travelling Salesman Problem (GTSP) \cite{ref:suarez2018robotsp}. 
Since finding the exact solution for a large-size GTSP is inapplicable in practice, we reduce the 
planning time by breaking it down into finding the target sequence that optimizes the Euclidean 
distance of the end-effector followed by optimizing the configuration path. 
To do so, we adopted the RoboTSP algorithm \cite{ref:suarez2018robotsp} with several modifications:

\subsubsection{Finding base sequence}
Find the optimal sequence of base poses $seq(\mathbf{b}(p+2))$ that minimizes the length of the base 
pose tour which starts from the home pose $\mathbf{b}_{home}$, travels through all $p$ base poses, 
and back to $\mathbf{b}_{home}$. 
The sequence is found as a solution to the Travelling Salesman Problem (TSP). 
For TSP solver, we use the 2-Opt algorithm \cite{ref:applegate2011traveling} as it yields high-quality 
solutions in low computation time \cite{ref:suarez2018robotsp}. 

\subsubsection{Finding target sequence}
Find the length-optimal target sequence $seq(\mathbf{t}(n+2))$ in task-space given $seq(\mathbf{b}(p+2))$. 
The tour starts from a home end-effector's pose $\mathbf{t}_{home}$, visits $n$ targets 
$\mathbf{t}_1, \dots, \mathbf{t}_n$, and back to $\mathbf{t}_{home}$. 
We propose solving a TSP on all targets while respecting the base sequence by virtually separating 
the clusters by a distance $h$ larger than the longest distance between targets in any cluster: 
\begin{equation}
        h = h_{scale} \cdot \max_{1 \leq k \leq p} \\
        \{\max_{i,j \in C_k} \\ 
        \{ \|\mathbf{r}_i-\mathbf{r}_j\| \} \} \\
\end{equation}
where $\mathbf{r}_i = (x^t_i, y^t_i, z^t_i)$ is the position of target $i$, and $h_{scale} \geq 1$ 
is a tuning parameter ($h_{scale} = 1$ in our experiments). 

In comparison with finding length-optimal target sequence for each cluster individually, our approach 
minimizes the total task-space path length of the end-effector including transitions between clusters. 
In terms of motion time, we have noticed that the former has higher time-optimality when the clusters 
are far apart while the latter produces faster paths when clusters are closer. 
We prefer the latter approach since a long-distance movement of the manipulator during base transitions 
leads to a large shift of the centre of gravity which may cause wheel's slip contributing to 
localization errors. 

\subsubsection{Finding configuration sequence}
Similar to RoboTSP \cite{ref:suarez2018robotsp}, we construct an undirected graph of $n+2$ layers 
following the sequence $seq(\mathbf{t}(n+2))$. 
Each layer in $[2,n+1]$ contains nodes representing IK solutions to reach a target, while the start 
and end layers represent home configuration. 
The shortest configuration path is found using a graph search. 

Finally, based on the base pose tour and manipulator configuration sequence, we can compute fast 
collision-free C-space trajectories using classical motion planning algorithms such as RRT-Connect 
with post-processing \cite{ref:kuffner2000rrt, ref:pham2015trajectory}.

\section{EXPERIMENTS}

\subsection{Real experiment}

\begin{figure}[tb]
        \centering
        \begin{subfigure}{0.35\linewidth}
                \includegraphics[width=1\linewidth]{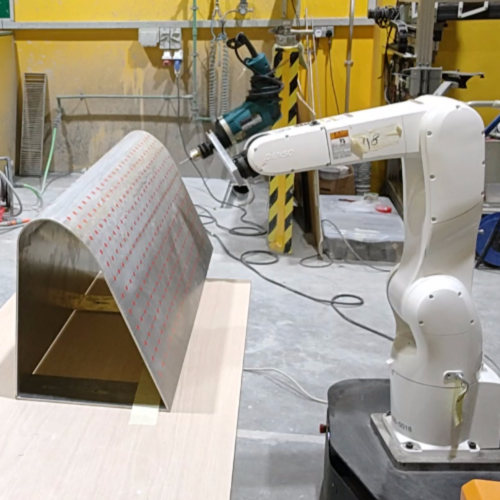}
                \caption{Visit the 1\textsuperscript{st} cluster.}
        \end{subfigure}
        \hspace{0.05\textwidth}
        \begin{subfigure}{0.35\linewidth}
                \includegraphics[width=1\linewidth]{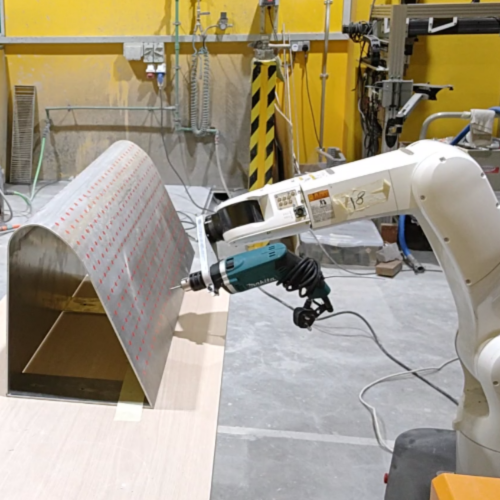}
                \caption{Visit the 2\textsuperscript{nd} cluster.}
        \end{subfigure}
        \hfill
        \begin{subfigure}{0.35\linewidth}
                \includegraphics[width=1\linewidth]{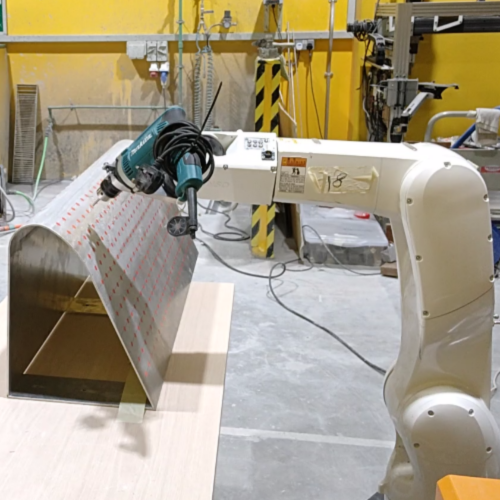}
                \caption{Visit the 3\textsuperscript{rd} cluster.}
        \end{subfigure}
        \hspace{0.05\textwidth}
        \begin{subfigure}{0.35\linewidth}
                \includegraphics[width=1\linewidth]{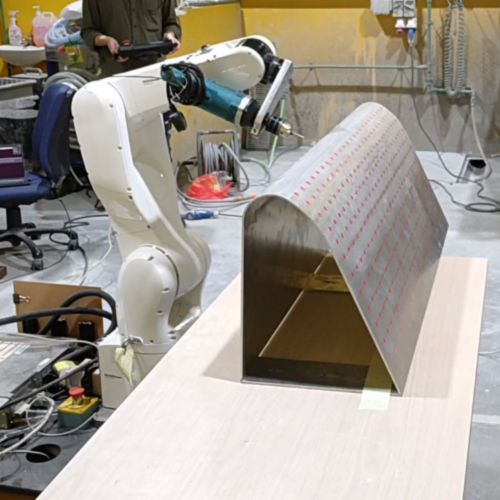}
                \caption{Visit the 4\textsuperscript{th} cluster.}
        \end{subfigure}
        \caption{Real experiment of the mobile drilling task in Fig. \ref{fig:demo_task}.}
        \label{fig:demo_sol}
\end{figure}

We implemented our method in a task and motion planning algorithm for a mobile drilling task 
containing 336 targets with a wide range of locations and orientations (Fig. \ref{fig:demo_task}). 
Our mobile manipulator consists of DENSO VS-087 6-DOF manipulator mounted on Clearpath Ridgeback 
3-DOF omnidirectional mobile base. 
An OpenRAVE simulation environment \cite{ref:diankov2010automated} was used to mimic and solve the 
task, using LRg solver, $0.10m$ floor grid size, and $0.05m$ voxel size (explained in VI.B). 
Our algorithm found 4 clusters with 4 corresponding base poses (Fig. \ref{fig:solution}, 
\ref{fig:demo_sol}), motion sequence and collision-free trajectories to visit all targets, 
in 100 seconds. The total motion time was 414 seconds. 
The experiment is shown in an accompanying video (https://youtu.be/Vopupf81hYo). 

\subsection{Benchmarking}

We benchmarked the key elements of our proposed algorithm in a mobile drilling task similar to 
Fig. \ref{fig:demo_task}, with a wider azimuthal range of $[-46^\circ,46^\circ]$ (front) and 
$[152^\circ,208^\circ]$ (back), in OpenRAVE simulations. 
All simulations were performed on CPU AMD\textsuperscript{\textregistered} Ryzen 9 running Ubuntu 18.04.

\subsubsection{Benchmarking online processes}

Since the sequencing method is based on RoboTSP which was benchmarked in \cite{ref:suarez2018robotsp}, 
here we only evaluate the clustering method in more details. 
The benchmarking parameters include the number of targets, grid size of the floor, and SCP solvers 
such as: 
\begin{itemize}
        \item \textit{Greedy}: at each iteration, choose the set with the highest number of uncovered 
        elements \cite{ref:williamson2011design, ref:zhu2016new}. 
        \item \textit{Linear Programming relaxation and rounding (LPr)}: formulate SCP as Integer 
        Program (IP) and relax it into Linear Program, then apply deterministic rounding 
        \cite{ref:williamson2011design}. 
        \item \textit{Lagrangian Relaxation and greedy (LRg)}: we implemented the SetCoverPy algorithm 
        \cite{ref:zhu2016new} which relaxes the IP by adding a Lagrange multiplier vector to the cost 
        function, and combines with greedy algorithm. 
        The author of \cite{ref:zhu2016new} reported $99\%$ optimality on their tests. 
\end{itemize}

\begin{figure}[tb]
        \centering
        \begin{subfigure}{0.49\linewidth}
                \includegraphics[width=1\linewidth]{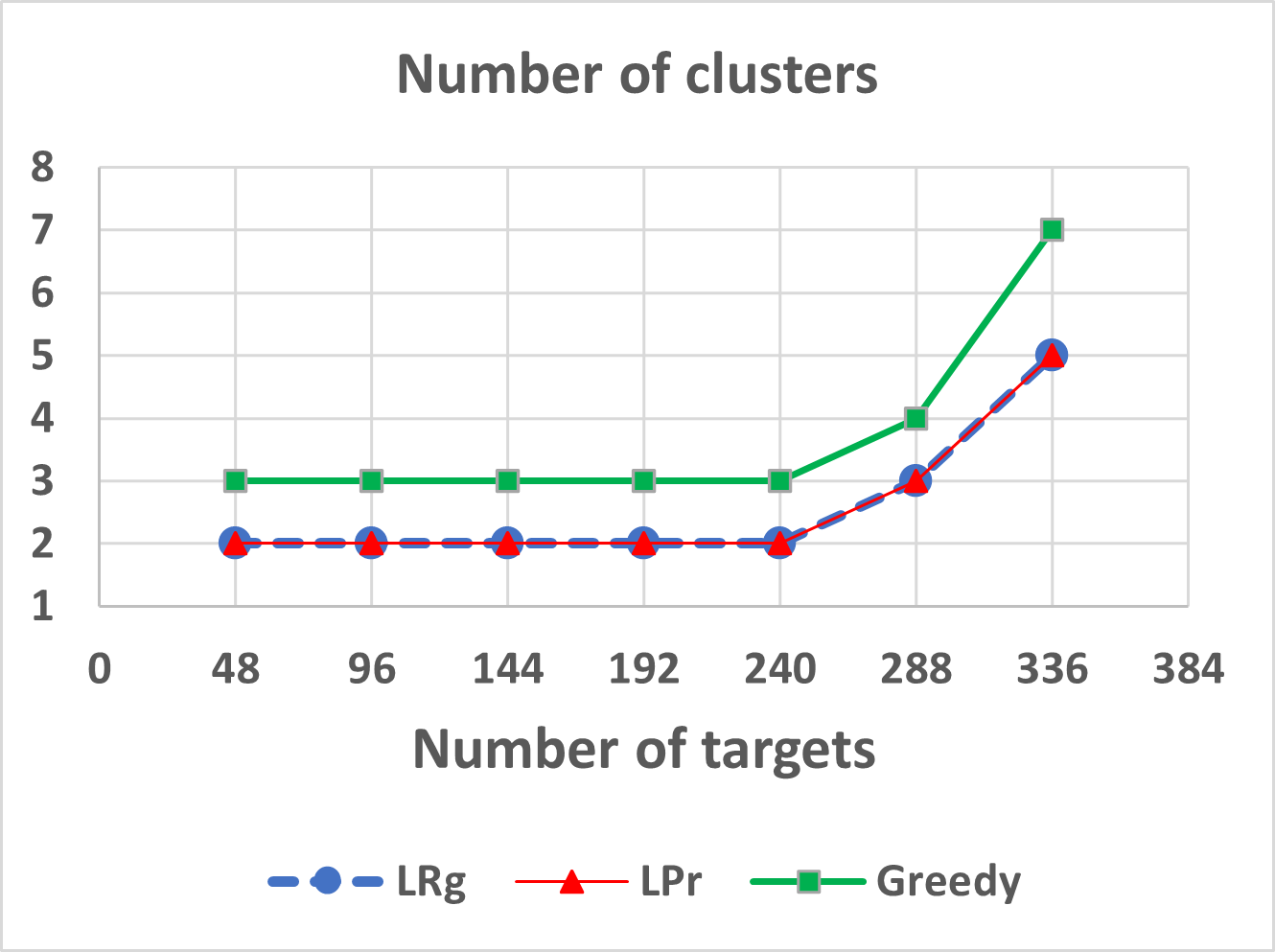}
                \caption{Floor grid size: $0.10m$.}
        \end{subfigure}
        \hfill
        \begin{subfigure}{0.49\linewidth}
                \includegraphics[width=1\linewidth]{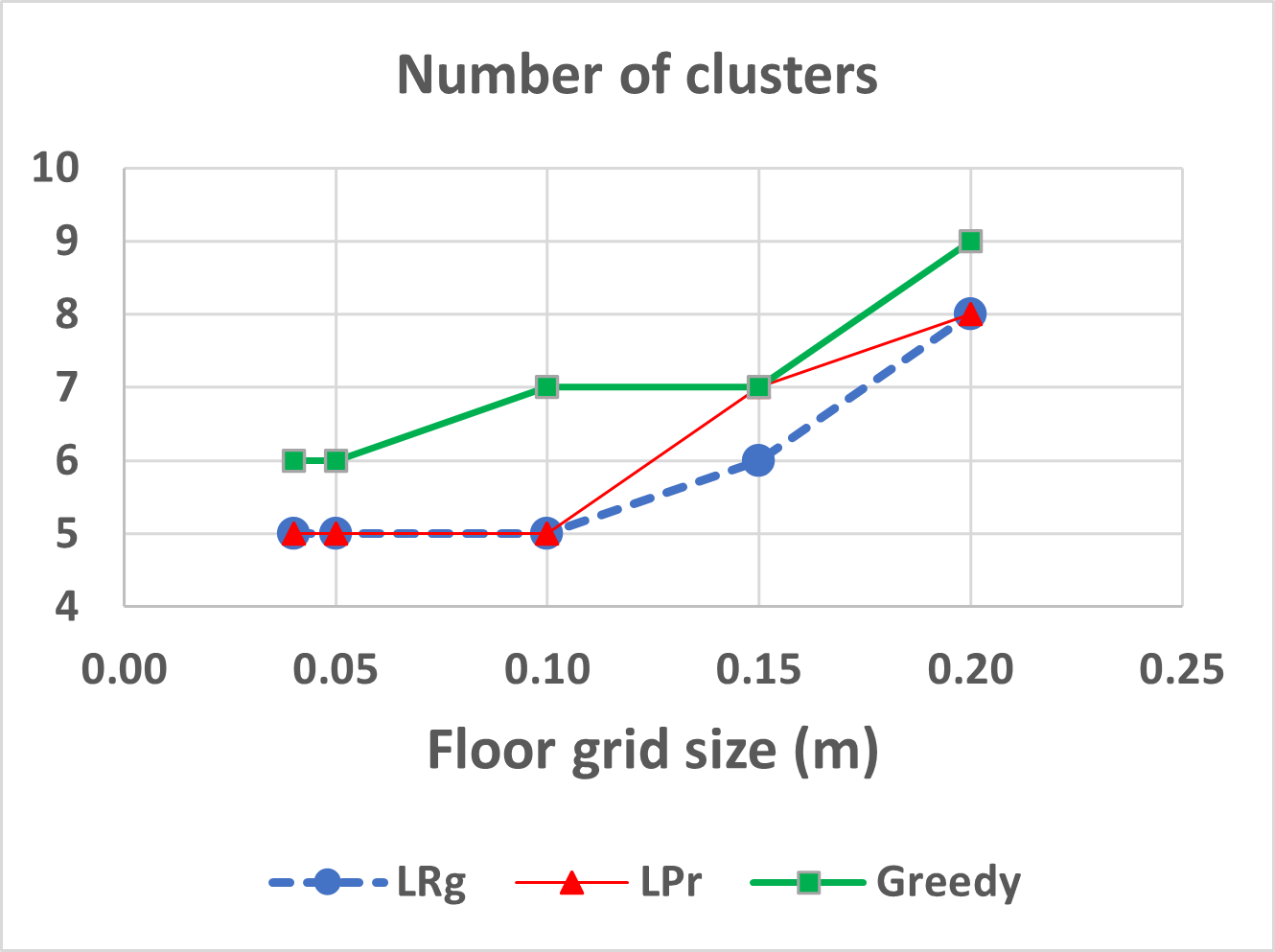}
                \caption{Number of targets: 336.}
        \end{subfigure}
        \hfill
        \begin{subfigure}{0.49\linewidth}
                \includegraphics[width=1\linewidth]{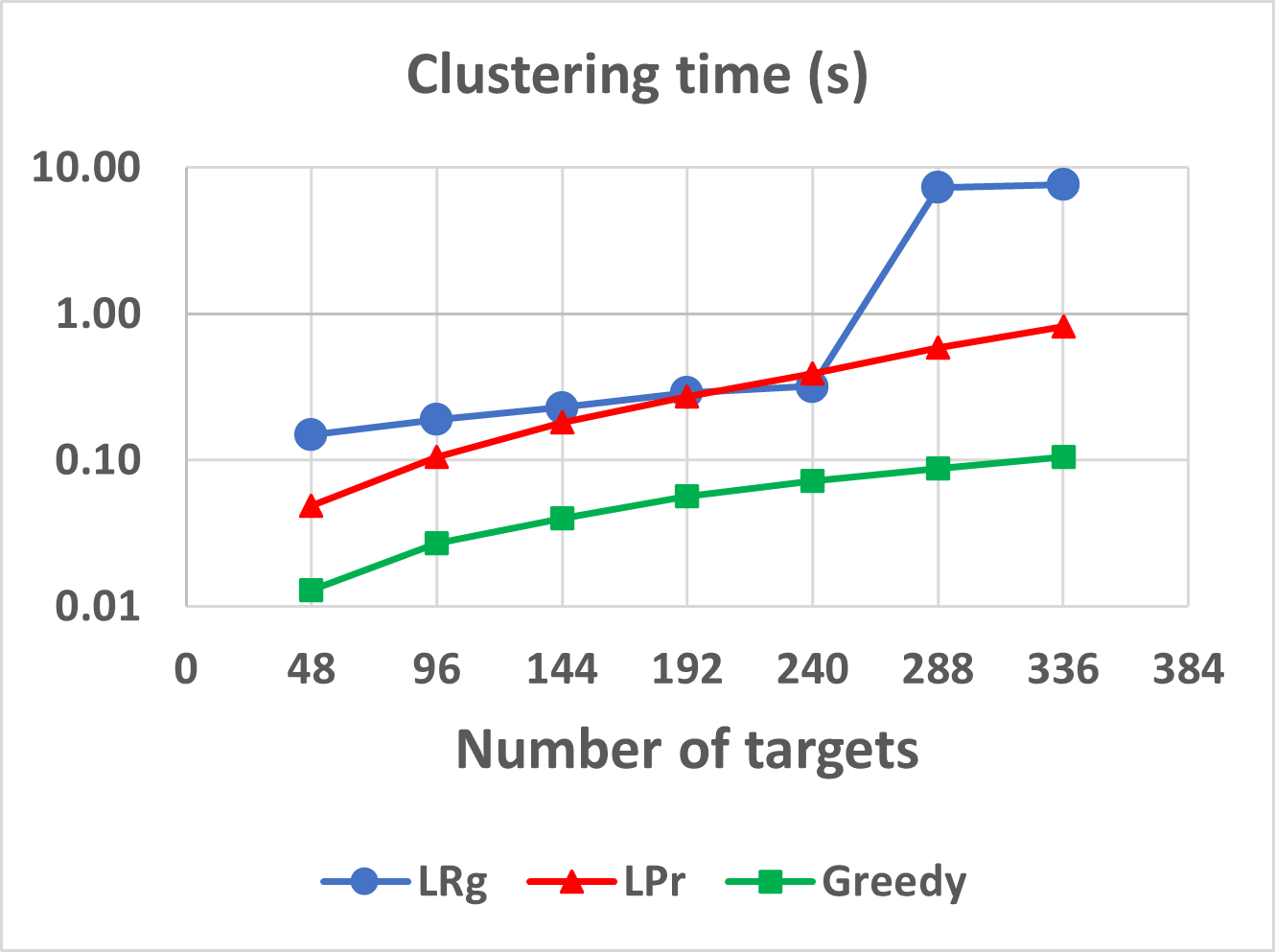}
                \caption{Floor grid size: $0.10m$.}
        \end{subfigure}
        \hfill
        \begin{subfigure}{0.49\linewidth}
                \includegraphics[width=1\linewidth]{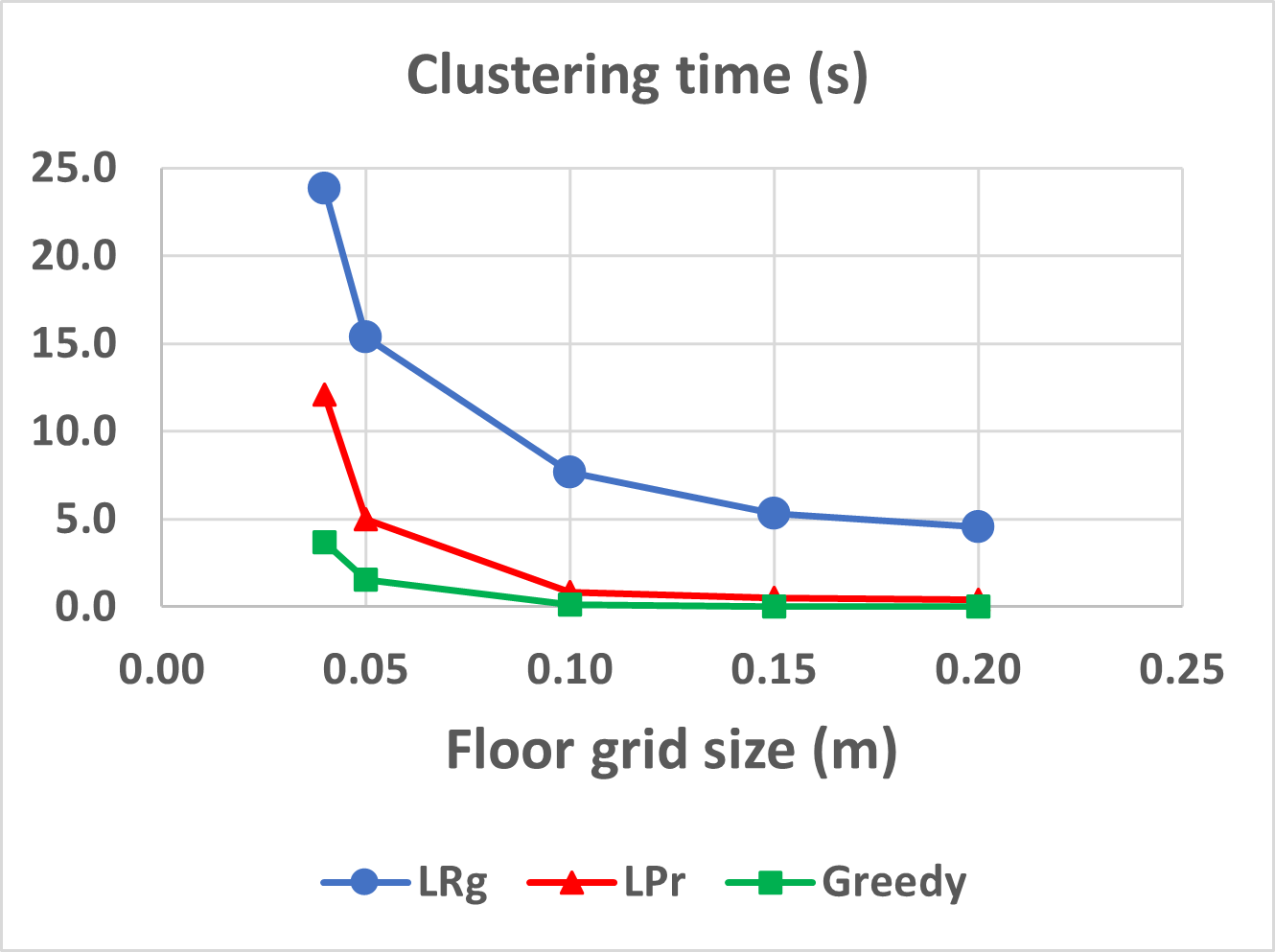}
                \caption{Number of targets: 336.}
        \end{subfigure}
        \caption{Benchmarking the clustering process.}
        \label{fig:benchmark}
\end{figure}

\begin{figure}[tb]
        \centering
        \includegraphics[width=0.57\linewidth]{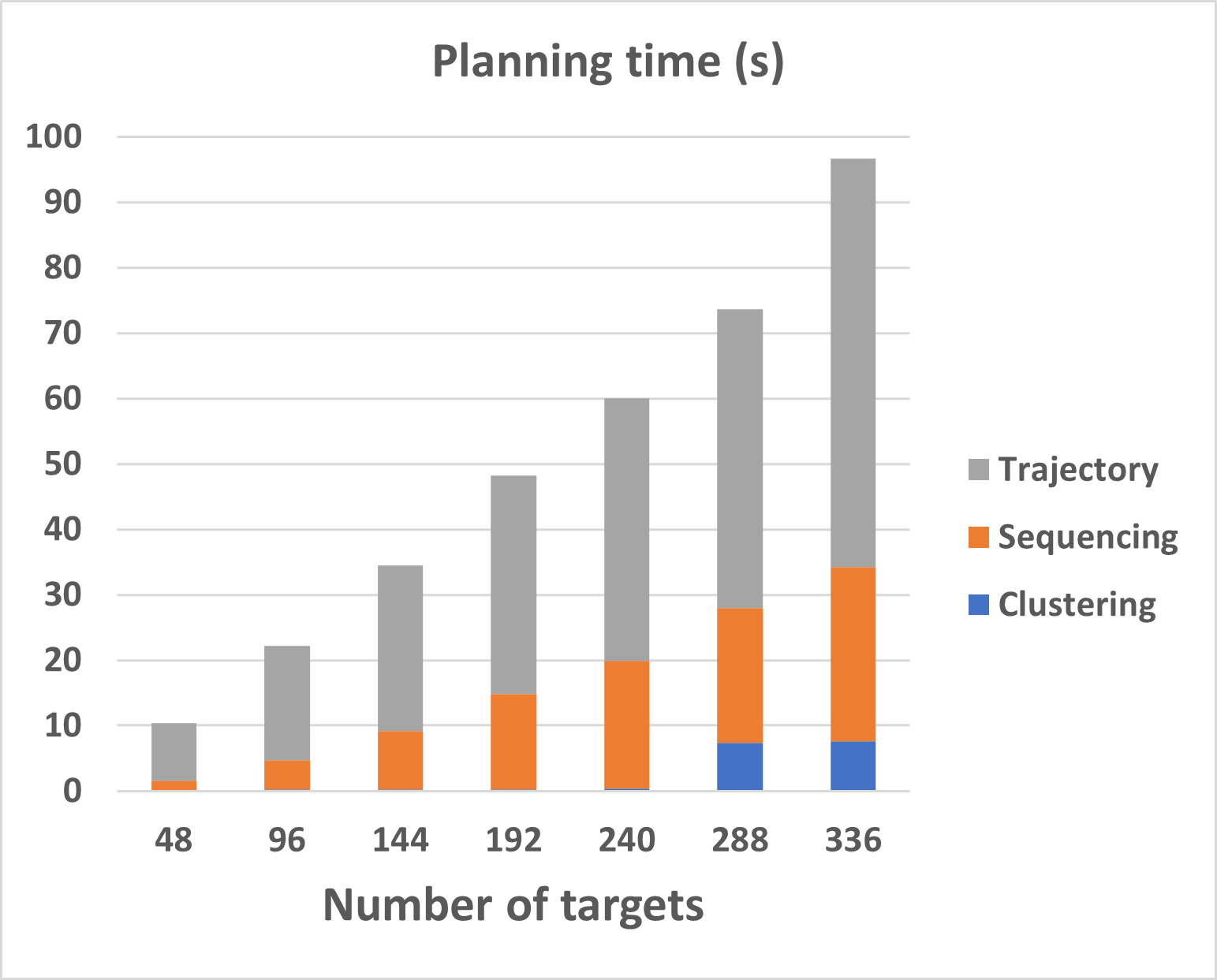}
        \caption{Planning time breakdown.}
        \label{fig:time_online}
\end{figure}

The evaluation metrics are the SCP solution (number of clusters found) and computation time. 
The benchmarking results are shown in Fig. \ref{fig:benchmark}. 
In most cases, LRg solver found the best solution and consumed longer but acceptable computation time. 
We used LRg solver, set grid size at $0.10m$, and recorded the computation time spent on different 
steps of the task and motion planning algorithm in Fig. \ref{fig:time_online}. 

\subsubsection{Benchmarking offline process}
We consider the reachability database generation as an offline process since it can be reused for 
tasks with target orientations within the sampling polar angle range, using the same robot. 
The computation time for different voxel size values is shown in Table \ref{tab:benchmark_fkr}. 

\begin{table}[h]
        \caption{Benchmarking reachability database generation.}
        \label{tab:benchmark_fkr}
        \begin{center}
        \begin{tabular}{|c|c|c|}
        \hline
        Voxel size (m) & Number of voxels & Computation time (s)\\
        \hline
        0.04 & 74004 & 1765.2\\
        0.05 & 40204 & 943.5\\
        0.07 & 15902 & 366.6\\
        0.10 & 6332 & 126.1\\
        \hline
        \end{tabular}
        \end{center}
\end{table}

\subsection{Comparison}

To the best of our understanding, there is no open-source algorithm that can solve our experiment and 
benchmark tasks where targets have a wide range of orientations. 
Therefore, in this subsection, we consider only 264 targets on one side of the workpiece with 
$110^\circ \leq \theta^t_i \leq 150^\circ, \phi^t_i = 0 \ \forall i$. 
We compare our algorithm with two state-of-the-art RTSP solvers: 
\begin{itemize}
        \item Baseline 1: RoboTSP extended with sphere-clustering. 
        Since RoboTSP \cite{ref:suarez2018robotsp} is designed for fixed-base robots, we add a 
        clustering step which clusters the targets into spheres that fit inside the cloud of valid 
        voxels. This sphere-clustering is then solved using greedy colouring of connected sequential 
        breadth-first search variant \cite{ref:kosowski2004classical}. 
        \item Baseline 2: Cluster-RTSP algorithm proposed in \cite{ref:wong2020novel}. 
\end{itemize}

The comparison results are shown in Table \ref{tab:comparison}. 
In this test case, our method appeared to be the fastest solver while produced the minimum number 
of base poses and shorter motion time as compared to two baseline methods. 
The simulations are shown in the accompanying video. 

\begin{table}[hb]
        \caption{Comparison with existing methods}
        \label{tab:comparison}
        \begin{center}
        \begin{tabular}{|c|c|c|c|}
        \hline
        Algorithm & Base poses & Planning time (s) & Motion time (s)\\
        \hline
        Our method & 2 & 49.7 & 225.4\\
        Baseline 1 & 8 & 51.8 & 265.0\\
        Baseline 2 & 70 & 368.1 & 323.5\\
        \hline
        \end{tabular}
        \end{center}
\end{table}

\section{CONCLUSION}

In this paper, we have proposed a fast, near-optimal task-space clustering method for solving the 
Mobile Manipulator RTSP. 
In a mobile drilling task where 336 targets were distributed on the surface of a workpiece and had a 
wide range of orientations, our algorithm divided all targets into 4 clusters with 4 corresponding 
base poses to reach them, then found length-optimal configuration sequence and collision-free 
trajectories to visit every target, in 100 seconds. 
This method can be used in practical applications such as drilling, spot-welding, surface finishing, 
inspection scanning, picking and placing, etc. 
Our future directions include extending the proposed method for mobile manipulators with the base 
moving in 3D space such as in gantry systems, and developing a clustering method for continuous tasks 
such as 3D printing.

\section*{ACKNOWLEDGMENT}

This research was supported by the National Research Foundation, Prime Minister's Office, 
Singapore under its Medium-Sized Centre funding scheme, CES\textunderscore SDC Pte Ltd, 
Sembcorp Architects \& Engineers Pte Ltd, and Chip Eng Seng Construction Ltd.

\addtolength{\textheight}{-11.55cm}   

\bibliographystyle{IEEEtran}
\bibliography{IEEEabrv, references}

\end{document}